# A Laplace diffusion-based transformer model for heart rate forecasting within daily activity context

Andrei Mateescu[1], Ioana Hadarau[1], Ionut Anghel*[1], Tudor Cioara[1], Ovidiu Anchidin[2] and Ancuta Nemes[2]

[1] Computer Science Department, Technical University of Cluj-Napoca, Memorandumului 28, 400114 Cluj-Napoca, Romania; andrei.mateescu@cs.utcluj.ro; hadarau.em.ioana@student.utcluj.ro; ionut.anghel@cs.utcluj.ro; tudor.cioara@cs.utcluj.ro.

[2] Niculae Stancioiu Heart Institute, Calea Moților 19-21, 400001 Cluj-Napoca, Romania; anchidin_ovidiu@institutulinimii.ro; nemes_ancuta@institutulinimii.ro.

*Corresponding author.

**ABSTRACT** With the advent of wearable Internet of Things (IoT) devices, remote patient monitoring (RPM) emerged as a promising solution for managing heart failure. However, the heart rate can fluctuate significantly due to various factors, and without correlating it to the patient's actual physical activity, it becomes difficult to assess whether changes are significant. Although Artificial Intelligence (AI) models may enhance the accuracy and contextual understanding of remote heart rate monitoring, the integration of activity data is still rarely addressed. In this paper, we propose a Transformer model combined with a Laplace diffusion technique to model heart rate fluctuations driven by physical activity of the patient. Unlike prior models that treat activity as secondary, our approach conditions the entire modeling process on activity context using specialized embeddings and attention mechanisms to prioritize activity specific historical patents. The model captures both long-term patterns and activity-specific heart rate dynamics by incorporating contextualized embeddings and dedicated encoder. The Transformer model was validated on a real-world dataset collected from 29 patients over a 4-month period. Experimental results show that our model outperforms current state-of-the-art methods, achieving a 43% reduction in mean absolute error compared to the considered baseline models. Moreover, the coefficient of determination $R^2$ is 0.97 indicating the model predicted heart rate is in strong agreement with actual heart rate values. These findings suggest that the proposed model is a practical and effective tool for supporting both healthcare providers and remote patient monitoring systems.

## 1. Introduction

Patients with heart failure are associated with an increased rate of mortality and high costs for the healthcare system. Therefore, attempts need to be made for identifying predictive factors of decompensation and hospitalisations. One promising solution is Internet of Things (IoT)-based remote patient monitoring (RPM) of health parameters to get objective insights in health progression [1]. In this context, heart rate (HR) is one of the most frequently analysed parameters. For patients with congestive heart failure with preserved left ventricular ejection fraction, it was observed that an increase of at least 5 beats per minute in resting HR compared to the previous visit is associated with an increased risk of cardiovascular mortality and hospitalisation [2]. An increase in the weekly average nocturnal HR by 5 beats per minute was associated with a more than twofold risk of cardiovascular mortality and rehospitalisation in patients with heart failure [3]. Rhythm disorders play an important role in the decompensation of patients with chronic heart failure. Approximately 30% of episodes of acute heart failure are precipitated by atrial fibrillation, 7% by ventricular arrhythmias, and 4% by bradycardia [4].

However, context-related challenges are an important and often overlooked issue in remote HR monitoring. These relate to interpreting HR data accurately considering what the person is doing, or experiencing, rather than assuming every fluctuation is clinically significant [5]. At the same time, patients are wearing devices that monitor their physical activity and HR at home, such as smartwatches. These patients often present the recorded data to the doctor at the time of consultation. Doctors recognise their potential, but they are unsure of the existing scientific evidence and do not always know how to use the monitored parameters [6]. Compared to continuous Holter electrocardiogram (ECG) monitoring that provides information over short periods of time, smart devices collect data related to the wearers' condition and activity over long periods of time, which could offer a better overview of their overall health.

Chest straps with electrodes for temporary HR monitoring have been used in sports for a long time. Devices worn on the wrist based on photoplethysmography continuously monitor the same parameters but are more convenient. The improvement of this technique has led to the incorporation of sensors into miniature devices such as rings or glasses [7,8]. However, the occurrence of rhythm disorders (extrasystoles, atrial fibrillation) can lead to an erroneous interpretation of the HR based on photoplethysmography, which can be partially corrected using complex algorithms. Ultimately, modern watches can counteract this shortcoming through direct ECG recording, which also has a software for simplistic automatic interpretation of HR and rhythm [9]. Moreover, wearable devices can monitor a variety of additional parameters such as oxygen saturation to step counts, minutes of activity, bringing contextual information which helps in interpreting HR more accurately. Of these, HR and physical activity are the easiest parameters to monitor in the general population with the help of these devices and the vast majority have these recording capabilities [10]. They should be recommended to patients with symptoms suggestive of rhythm or conduction disorders, or those already receiving medication to assess response to treatment [11].

Correlating activity information with recorded parameters can have an important role in making medical decisions. In general, this information is obtained via anamnesis, but in the case of older patients this could be difficult to do. Different wearable devices allow users to record the occurrence of symptoms, at which point they could be correlated with the simultaneously recorded parameters. Finally, these devices could help in the differential diagnosis of cardiac diseases from those of non-cardiac origin, thus helping identify patients who would require further investigations. Elevated HR due to running, might mimic tachycardia or other cardiac events and without activity context, such increases could trigger false alarms [12].

Artificial Intelligence (AI) models have great potential for increasing the accuracy, personalization, and context-awareness of remote HR monitoring [13, 14]. They may help in corelating the HR with activity, sleep or stress, to distinguish between normal and abnormal changes [15]. Moreover, such models can potentially learn an individual's baseline physiology, reduce false alarms and improve relevance [16]. However, modelling of HR during physical activities introduces its own set of technical challenges. Exiting models struggle to capture complex patterns that include gradual physiological trends and sudden HR transitions caused by activity changes. HR data collected from wearables are heterogeneous, noisy, have variable scales and resolutions, and present complex dependencies between datapoints, making them difficult to model. Physiological time series are affected by many factors, making them non-linear and nonstationary, and, therefore, HR time series are often considered difficult to predict and cannot be handled by classical machine learning or statistical models. Learning from different types of data with variability in sample rate, formats and missing data patterns requires complex fusion-based deep learning architectures [17]. Neural networks like Convolutional (CNN) or Recurrent (RNN) neural networks may struggle with long-term dependencies and often suffer from the vanishing gradient problem, limiting their effectiveness over long HR data sequences [18]. While Long Short-Term Memory (LSTM) models partially address this issue, they still struggle with very long sequences and complex temporal relationships between HR patterns and activity transitions [19]. The RNN model proposed in [20] incorporates multimodal activity features to successfully predict the next activity, but relies on data from multiple sensors (ECG, accelerometer, gyroscope, etc) and treats these features as generic inputs rather than explicitly modelling their relationships. Recent advances in transformer-based architectures have shown promise for addressing some of these challenges. Transformers use self-attention mechanisms to weigh the importance of different parts of the input data sequence, allowing them to capture relationships between distant data elements more effectively [21]. However, existing transformer applications to HR modelling have primarily focused on general prediction tasks without explicitly addressing the activity-conditioning aspect. Additionally, these models require significant amounts of data and existing data sets often lacks annotated activity related context (e.g., what the patient was doing at the time). Wearable device limitations add to these challenges, as the measurements have a higher error rate during activity than at rest [22]. The integration of activity context into HR modelling remains a notable challenge, as current approaches often treat activity information as auxiliary features rather than main components of the model. This limitation becomes problematic when modelling sudden HR transitions that occur during activity changes, such as transition from rest to exercise or between different exercise intensities. Current diffusion models [23] generate smooth transitions but underestimate HR volatility during intense activity and can miss sharp transitions of the signal. The inherent Gaussian noise in these models fails to capture the heavy-tailed nature of physiological signals during rapid activity changes, leading to suboptimal reconstruction of sudden HR spikes or drops that are characteristic of exercise transitions.

Our work addresses these limitations by introducing an activity-conditioned Transformer along with a Laplace diffusion model, explicitly designed to model activity-driven HR fluctuations. Unlike previous approaches that treat activity as secondary information, our method conditions the entire modeling process on activity context. We build upon the vanilla Transformer architecture, chosen due to its ability to capture long-term dependencies between the data points of the sequence through attention layers, and adapt it to model the complex relationship between HR and physical activity. First, we introduce activity-contextualized embeddings that encode both categoric and continuous data, including the type and intensity level of physical activities, HR derived feature and temporal information. These embeddings condition the Transformer's self-attention mechanisms to prioritize activity-specific historical patterns. Second, we design activity-specific encoder layers that process these embeddings through dedicated attention heads, enabling the model to capture the distinct physiological characteristics of different activities. This approach goes beyond generic HR modeling by explicitly learning activity-specific patterns. Third, we integrate a diffusion process that utilizes Laplace noise to address the challenge of modeling HR variations due to activities changes. The heavy-tailed Laplace distribution models sudden HR transitions during rapid activity intensity changes or between different exercise stages. The noise sampled from Laplace distributions allows for more outliers, enabling the model to reconstruct those sharp transitions during the reverse diffusion process. These adaptations address the main challenge of forecasting HR during physical activity, where conventional models fail to capture both long-term trends and sudden activity-driven transitions.

The main contributions of the paper are the following:

- Activity-contextualized embeddings that aggregate categorical (activity type), ordinal (intensity level, temporal data), and continuous (HR-derived features) inputs into a single representation, enabling the transformer model to prioritize activity-specific historical HR patterns.
- Activity-specific encoder layers, where dedicated attention heads learn HR patterns unique to different exercises, incorporating activities as main components of the model rather than auxiliary features.
- A Laplace diffusion model that captures fast HR transitions due to the heavy-tail properties of the Laplace distribution, allowing the transformer model to reconstruct HR variations during activities changes or recovery phases.

The rest of the paper is structured as follows: Section 2 presents the state of the art on transformer models in healthcare scenarios, Section 3 describes the transformer model, Section 4 presents and discusses the relevant evaluation results in the context of a real-world dataset, while Section 5 presents conclusions and future work.

## 2. Related work

The integration of deep learning models and probabilistic forecasting techniques into IoT and RPM technologies with the objective of generating insights for doctors from historical patient data has generated many research approaches in the last years. Traditional time series models like Autoregressive integrated moving average (ARIMA) and Prophet were commonly used in the past for forecasting, but they struggle with complex datasets as the one in the healthcare domain are.

In contrast, modern deep learning methods, capture complex nonlinear patterns and long-term dependencies in HR data that traditional models often miss, thus delivering superior predictive performance [19, 24]. Ghafoori et al. [24] evaluated multiple deep learning models, including LSTM, Gated recurrent units (GRU) and CNN-based architectures for predicting construction workers' HR. The models achieved Mean Absolute Error (MAE) values between 5.40 and 6.72 when forecasting one-minute-ahead values. Similarly, Ni et al. [25] evaluates traditional time series models and compares them with modern deep learning approaches for HR prediction. The study highlights the limitations of classic models like ARIMA, which performs well in modeling seasonal variations, but struggles with noisy and sparse datasets, and Prophet, which requires accurate hyperparameter tuning and is less suited for real-time data. Zhu et al. [26] train multiple LSTM models to forecast HR values during different activities. They achieve high accuracy for 5 seconds ahead forecasting, but their models struggle with longer horizons. Deep learning models consistently outperformed traditional statistical models, but the new transformer-based architectures have been adapted for the specific case of time-series analysis and achieved best overall results [27]. Consequently, multiple recent studies have demonstrated that transformer-based architecture provides significantly better results by capturing long-term dependencies and context [27-29].

Recent studies have shown that Transformer-based architecture is driving significant innovations across multiple healthcare sectors [30]. The application of such models for human activity recognition (HAR) has been explored over the last couple of years, as noted by [31]. The models proposed were adapted to classify activities using smartphone motion sensor data from accelerometers and gyroscopes. The transition from traditional pretrained language models to large language models (LLMs) in the healthcare domain is described in [32]. The authors highlight the ability of LLMs to handle complex clinical tasks like relation extraction and medical document classification and discuss several state-of-the-art models used in healthcare. They emphasize the importance of fine-tuning models for specialized medical tasks and address ethical and security-related challenges. In [33] the authors introduce a Two-Stream Transformer designed for HAR using wearable sensor data. The paper addresses the challenges of capturing spatial-temporal dependencies and handling data coming from sensors positioned at different body locations which contribute differently to the classification process. To address these issues, the authors propose an architecture with a temporal stream used for extracting sequential features and a spatial stream used for learning relationships between multiple sensors. Authors from [34] propose a Transformer-based model for processing unstructured clinical notes from electronic health records (EHRs) to improve hospital readmission prediction. They fine-tune the Bidirectional Encoder Representations from Transformers (BERT) model by assigning a dynamic risk score based on patient notes, making it usable by medical staff for early intervention. Other models generate detailed medical image reports by using Dense Convolutional Network (DenseNet) based attention for region detection with parallel, non-recurrent decoders, thereby overcoming the limitations of traditional RNN approaches [35]. In affective computing, researchers have developed models that fuse multi-modal physiological signals, such as skin conductance, and electroencephalogram (EEG), using self-attention mechanisms. These mechanisms capture both long-term dependencies and cross-modal interactions, which improves the recognition of affective states in everyday environments [36]. Moreover, models like BEHRT and Med-BERT have extended the Transformer framework to electronic health records. They learn contextualized embeddings of diagnoses by incorporating extra information such as the order of visits and patient age [37, 38]. These methods not only improve the accuracy of disease prediction but also enhance interpretability through attention visualization. Collectively, these studies highlight the promise of Transformer architectures for healthcare applications, extending medical imaging and physiological signal analysis to the complex analysis of longitudinal EHR data by utilizing parallel processing, modelling long-range dependencies, and generating rich, contextual feature representations. In [39] the authors propose a model that utilizes Transformers' self-attention mechanism to improve predictions of patient heath status, achieving better results than models like LSTM and CNN. The model was tested on multiple tasks such as predicting vital sign deterioration and respiratory rate over future timestamps, estimating mortality as a binary classification task and predicting the patient's remaining length of stay in the hospital.

Transformer models can be successfully used for clinical time series analysis and prediction for the specific case of HR disease management as noted in [25]. For cardiovascular diseases (CVD), a multi-modal approach for estimating heart failure is explored in [40] to address the challenges of traditional risk assessments which rely on costly diagnostic methods. The authors integrate ECG and HR values to estimate heart failure hospitalization risk. They propose a Transformer-enhanced ResNet Model which outperforms the other baseline models. A hybrid CNN-Transformer architecture is introduced by [41] to extract HR patterns in sparse datasets. They utilize transfer learning and fine tune on few participants to achieve better results than CNNs and other baselines. The study highlighted in [42] focuses on heart disease prediction using a Transformer-based model that learns from multiple patient features such as blood pressure, cholesterol levels and HR. The research notes the importance of feature selection and preprocessing to improve classification performance. Huang et al. [43] proposed a TE-SAGRU model that employs parallel transformer encoders combined with stacked attention gated recurrent units to integrate heterogeneous signals, such as PPG and ECG, thereby capturing the long-term dependencies, important for accurate continuous blood pressure monitoring. Shen [44] introduced TransformHR, employing self-attention mechanisms in a transformer framework to forecast HR from wearable devices during high intensity activities, thereby addressing challenges related to motion artifacts and energy constraints. The model achieves a MAE of 4.0-4.2 while being optimized for low computational cost. In the clinical risk assessment field, Antikainen et al. [45] used transformer models such as BERT and XLNet on diverse electronic health records to predict six-month mortality in cardiac patients, using the deep understanding of bidirectional context of BERT and XLNet's permutation-based training, which together improve the ability of the models to capture nuanced temporal dependencies and complex feature interactions. Meanwhile, Rao et al. [46] introduced an explainable transformer-based model for incident heart failure that integrates multimodal inputs with post-hoc perturbation analyses, not only clarifying critical risk factors but also providing valuable insights into feature importance and model decision pathways. Complementing this progress, Houssein et al. [47] effectively adapted pre-trained transformer language models for extracting heart disease risk factors from clinical narratives. These approaches demonstrate that transformer models, with their powerful self-attention mechanisms and multimodal integration capabilities, are redefining the landscape of cardiovascular health monitoring and risk prediction.

Recent advances in generative modeling have demonstrated strong performance of diffusion models, particularly denoising diffusion probabilistic models (DDPMs) and their variants (e.g. DDIMs), in time series forecasting [48], anomaly detection [49], synthetic data generation [50] and healthcare applications [51]. These models capture complex, multimodal data distributions through iterative denoising, achieving superior mode coverage compared to traditional generative approaches (e.g. Generative adversarial network - GAN, Variational Autoencoder - VAE) [52] by avoiding mode collapse and generating diverse samples that reflect variable patterns of the training data. This property is especially important for healthcare applications, where physiological signals like HR present different modes (e.g., resting, exercise, recovery) with sharp transitions between them. For instance, Tashiro et al. [23] handle the time series imputation task through a conditional score-based diffusion model (CSDI), which utilizes observed data to guide the denoising process, along with transformer networks, and achieves state-of-the-art performance in healthcare applications. Jenkins et al. [53] further demonstrated the potential of diffusion models for physiological signal processing by proposing a template-guided DDPM for ECG imputation, which addresses challenges such as missing data and morphological variability. The proposed model utilizes data-driven prior adapted to individual subjects and augmented with beat-level shifts and matches state-of-the-art performance for extended gaps. Moreover, diffusion models excel at maintaining inter-feature relationships and generating contextually correct interpolations [54], suggesting their potential to model smooth transitions between HR states. A Transformer-based Diffusion Probabilistic Model for Sparse Time Series Forecasting (TDSTF) model is proposed by Chang et al. [55] to address the challenges of sparse and irregularly sampled Intensive Care Unit (ICU) data. The approach is used to forecast HR, systolic blood pressure, and diastolic blood pressure in ICUs. The proposed model combines transformers with diffusion probabilistic models to improve the forecasting process. A triplet representation of data is employed to efficiently handle missing values in time series data. TDSTF outperforms previous models with better Mean Squared Error (MSE) and Standardized Average Continuous Ranked Probability Score (SACRPS) scores and faster inference speed.

The reviewed literature presents several limitations in case of HR modeling within daily activity context. Existing models struggle to capture complex patterns found in these time series, due to the combination of gradual changing trends of HR with sudden transitions caused by activity changes. HR data collected from wearables presents additional challenges due to its heterogeneous and noisy nature with variable scales and resolutions. Classical AI approaches are not suited for handling the non-linear characteristics of these time series, while neural network approaches like RNNs and CNNs suffer from the vanishing gradient problem and struggle with long-term dependencies and activities context integration. Although recent Transformer-based architectures can capture relationships between distant data points, existing applications have primarily focused on general prediction tasks without explicitly addressing activity contextualized HR prediction. Furthermore, state of the art considered diffusion models may generate unrealistic transitions and underestimate HR volatility during activity transitions. Therefore, the effective integration of specific activity context into HR prediction remains a challenge that the current state-of-the-art does not address

## 3. Contextualized HR transformer model

Accurate HR forecasting within daily activity context faces three main challenges:

- multimodal data integration, where continuous values derived from raw HR signals, categorical activity types and ordinal intensity levels must be processed, creates embedding conflicts;
- HR presents complex, non-stationary patterns that combine gradual trends with abrupt transitions during activity changes, requiring activity-aware modeling approaches;
- heavy-tailed distribution of HR values during exercise transitions is not properly captured by traditional models.

To address these challenges, we extend the vanilla Transformer architecture by incorporating domain-specific adaptations at multiple levels (see Figure 1). Section 3.1 presents our multimodal data fusion approach that designs context-aware embeddings which project continuous HR-derived signals (via 1D convolutions), temporal information extracted from timestamps and categorical activity labels into a single vector representation, preventing feature dominance through adaptive embeddings. These representations are then introduced into the Transformer model using a sliding window approach to enable continuous prediction. Section 3.2 addresses the challenge of complex, activity-dependent physiological patterns by introducing specialized encoders for each exercise type that enable the model to focus on relevant patterns specific to that activity context. After low-level processing, a redirection mechanism routes each input sequence to the appropriate activity encoder. Section 3.3 focuses on modeling the volatile HR dynamics during activity transitions by implementing a Laplace diffusion process where the Transformer network iteratively predicts and removes noise from target HR sequences. The heavy-tailed Laplace distribution is better suited for modeling sudden HR jumps and preserving recovery patterns post-exercise.

We train the model using the L1 loss, which is equivalent to minimizing the Laplace negative log-likelihood. During training, a small amount of noise corresponding to the noise schedule is added to HR data. The Transformer model is trained to predict this noise, updating its weights via L1 loss. During inference, the trained Transformer is utilized to remove noise step by step, starting from a sequence of full noise.

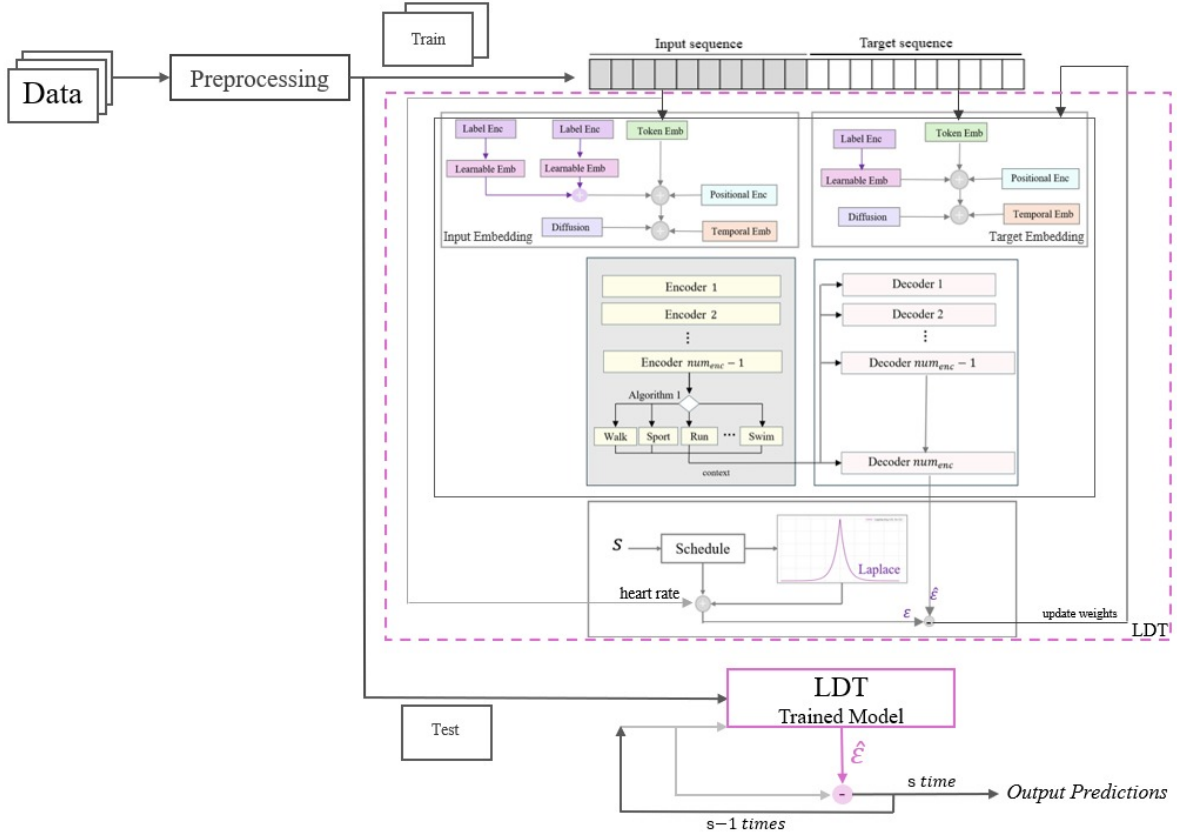

**Figure 1.** Transformer Model Architecture.

## 3.1. Multimodal data fusion

### 3.1.1. Temporal data and derived features integration

To predict and correlate HR in the context of physical activities, we represent multiple interrelated data series as embeddings, enabling fusion and modeling of complex physiological patterns. The HR time series for each patient $i$ is defined as a function that maps time points to HR values:

$$H_i: T \to \mathbb{R}^1 \quad (1)$$

where $T$ is a countable set of time points sampled at interval $\Delta t$ of 1-minute between observations. The sequence of HR data points $h_i(t)$ is indexed in time order:

$$H_i = \{ h_i(t_h), t_h \in T, \Delta t_H = t_{h+1} - t_h \} \quad (2)$$

The patient activity intensity levels $L_i$ is a multivariate time series that maps each time point to all the predefined activity intensity features in set $C$:

$$L_i: T \to \mathbb{R}^{|C|} \quad (3)$$

where $|C|$ is the set cardinality and the activities intensity features set is defined as:

$$C = \{c_n | c_n \in \{\text{sedentary, lightly\_active, fairly\_active, very\_active}\}\} \quad (4)$$

Therefore, the time series is formally represented as:

$$L_i = \{ l_i^C(t_l), t_l \in T, \ \Delta t_L = t_{l+1} - t_l \} \quad (5)$$

where the interval between sample observations $\Delta t_L$ is 1-minute and each element is a vector of values for each activity intensity feature:

$$l_i(t) = [l_i^{sedentary}(t), l_i^{lightly\_active}(t), l_i^{fairly\_active}(t), l_i^{very\_active}(t)] \tag{6}$$

We define a merged multivariate time series $Z_i$ as a fusion of the above series, represented by a mapping function $z_i$ over synchronized time point $t \in T$:

$$Z_i: T \to \mathbb{R}^{1+|C|}, z_i(t) = [h_i(t), l_i(t)] \tag{7}$$

with the time series represented as:

$$Z_i = \{z_i(t_z), t_z \in T, \Delta t_z = t_{z+1} - t_z\} \tag{8}$$

where the interval between sample observations $\Delta t_z$ is 1-minute and represents the complete multivariate time series for patient $i$, where each element corresponds to a synchronized observation combining HR and activity intensity levels into a single vector. This fusion is necessary for understanding the effects of activity intensity on the HR variations.

The time series are further contextualized by annotating with the specific physical activity label performed at each timestamp. The set of possible physical activities labels tracked by the wearable device is:

$$A = \{running, walking, swimming, aerobic\ workout, outdoor\ biking, sport, treadmill\} \tag{9}$$

For each patient $i$, a monitored activity segment $a_i^k$ is represented by a triplet containing the activity label $a \in A$, the start time $\tau \in T$ and the duration of the activity $\delta \in \mathbb{N}$, expressed in minutes:

$$\varepsilon_i = \{a_i^k = (a, \tau_i^k, \delta_i^k) | 1 \le k \le K_i\} \tag{10}$$

where the activity label is associated with the k$^{th}$ segment, $\tau_i^k$ is the start time of the segment, $\delta_i^k$ is the duration of the segment in minutes. $K_i$ is the number of total activity segments monitored for patient $i$ and is constrained by:

$$\tau_i^{k+1} \ge \tau_i^k + \delta_i^k, \forall\ 1 \le k \le K_i \tag{11}$$

meaning no two activity segments of a single patient overlap in time.

We define a time series labeling function $\varphi_i: Z_i \to A$, which assigns an activity label to each observation $z_i(t) \in Z_i$ as:

$$\varphi_i(z_i(t)) = a_i^k\ for\ t_{z_i} \in [\tau_i^k, \tau_i^k + \delta_i^k] \tag{12}$$

The label corresponds to the monitored activity segment $a_i^k$ if the timestamp of the observation falls within the interval of the activity segment. Therefore, the annotated series is:

$$\hat{Z}_i: T \to \mathbb{R}^{2+|C|} \tag{13}$$

$$\hat{Z}_i(t) = \{[z_i(t), \varphi_i((z_i(t)))]\ |\ t \in T, (\exists)k\ such\ that\ t_{z_i} \in [\tau_i^k, \tau_i^k + \delta_i^k]\} \tag{14}$$

To capture time-related patterns, we extract temporal features from time point $t$ such as the month of the year $m(t)$, day of the month $d(t)$, day of the week $w(t)$, hour of the day $h(t)$, and minute of the hour $n(t)$. Each feature is derived using a dedicated function $v$ that maps each timestamp $t \in T$ to a natural number with value ranges determined by the standard calendar and time system.

$$v: T \to \mathbb{N}^5,\ v(t) = [m(t), d(t), w(t), o(t), n(t)] \tag{15}$$

The multi-variate time series representation facilitates the comparison of time points by capturing temporal similarity, and enables models to incorporate recurring temporal patterns, such as daily and weekly cycles, that may influence variations in HR and activity behavior.

Additionally, to characterize HR dynamics by capturing short- and medium-term variations, we have considered several features derived from raw HR signals. We define the rate of change as a function $g_i$ that captures the difference between the current and previous HR values:

$$g_i: T \to \mathbb{R}^1, \quad g_i(t_h) = \frac{h_i(t_h) - h_i(t_h - 1)}{\Delta t_h} \tag{16}$$

To identify subtle fluctuations that may indicate transitions in physical activity or physiological state we compute the rolling standard deviation $r_i^{(5)}$ of the HR signal for patient $i$ at time $t_h$, computed over a fixed-size window of the previous 5-time steps:

$$r_i^{(5)}: T \to \mathbb{R}^1, \quad r_i^{(5)}(t_h) = \sqrt{\frac{1}{5}\sum_{j=0}^{5-1}(h_i(t_h - j) - \overline{h_i}^{(5)}(t_h))^2} \tag{17}$$

where the rolling mean $\overline{h_i}^{(5)}(t_h)$ is defined as:

$$\overline{h_i}^{(5)}(t) = \frac{1}{N}\sum_{j=0}^{5-1} h_i(t - j) \tag{18}$$

We computed the Exponential Moving Average $e_i$ for a time window of 5 to reveal the underlying trend in HR by filtering out high-frequency noise:

$$e_i^{(5)}: T \to \mathbb{R}^1, \quad e_i^{(5)}(t_h) = \alpha \cdot h_i(t_h) + (1 - \alpha) \cdot e_i^{(5)}(t_h - 1) \tag{19}$$

where $h_i(t_h)$ is the HR value and $\alpha$ is a smoothing factor. As it weighs recent values more heavily, it is useful for capturing the most current changes in HR behaviour.

Finally, we have determined smoothed N-minute HR trend $sm_i^{(N,M)}$ by computing the net change in HR over the past N minutes and then applies a rolling mean over the most recent M time steps to smooth out short-term noise:

$$sm_i^{(N,M)}: T \to \mathbb{R}^1, \quad sm_i^{(N,M)}(t_h) = \frac{1}{M}\sum_{j=0}^{M-1} p_i^{(N)}(t_h - j) \tag{20}$$

where the raw trend is first defined as:

$$p_i^{(N)}(t_h) = h_i(t_h) - h_i(t_h - N) \tag{21}$$

This feature highlights sustained trends in HR dynamics and detects transitions in activity, recovery phases, or abnormal fluctuations.

Therefore, a comprehensive feature vector that captures both the temporal context and the dynamics of the HR combines the previously defined components as:

$$f_i: T \to \mathbb{R}^{2+|C|+9}, \quad f_i(t) = [h_i(t), l_i(t), \varphi_i(z_i(t)), v(t), g_i(t), r_i^{(5)}(t), e_i^{(5)}(t), sm_i^{(5,15)}(t)] \tag{22}$$

To enable the Transformer model to learn temporal dependencies and transitions in HR data over fixed time intervals, we apply a sliding window approach with a window size of $L$ for both the input sequence and the target sequence. The sliding window is defined around the start of the activity segment $a_i^k$, with the $L$ steps prior to the start used as input, and the $L$ steps after the start used as output:

$$W_i^{source}(t) = [f_i(t - L), f_i(t - (L - 1)), f_i(t - (L - 2)) \ldots, f_i(t - 1)] \tag{23}$$

$$W_i^{target}(t) = [f'_i(t), f'_i(t + 1), f'_i(t + 2), f_i'(t + L - 1)] \tag{24}$$

where $t$ is the start timeslot of activity $a_i^k$. Equations (23) and (24) present the initial input and corresponding target sequence for each activity $a_i^k$ starting at time t. Subsequent windows are constructed using a non-overlapping sliding window approach. For each window, the last target sequence from the previous step is used as the input for the next step. We continue this process if there are at least $2L$ elements remaining, so the last window can be constructed without truncation. As a result, the model receives pre-activity physiological signals that often indicate upcoming transitions in activity levels, learning not only baseline HR patterns but also the physiological changes that preceded the start of physical activity.

### 3.1.2. Encoding and embedding

Figure 2 shows the encoding and embedding pipeline used for HR dynamics in the context of physical activity. Each element of the sliding window is a token corresponding to a specific value from one of the features at a given time step. We construct a single vector of length $5L$, consisting of five consecutive segments: HR, gradient, rolling standard deviation, exponential moving average, and HR trend, each contributing with $L$ values. The tokens are subsequently transformed through a combination of embeddings methods which are then integrated into a unified tensor representation and passed to the attention mechanism for downstream processing. We use a token embedding layer which generates an embedding vector of shape ($d_{model}, L$) applying 1D convolutions with a kernel size of 3 to enable the model to consider the previous, current, and next elements in the sequence for each feature. A circular convolution is used to handle boundary conditions, meaning that to compute the embedding for the first element, the model considers the last, first, and second values, thus maintaining continuity across the feature sequence.

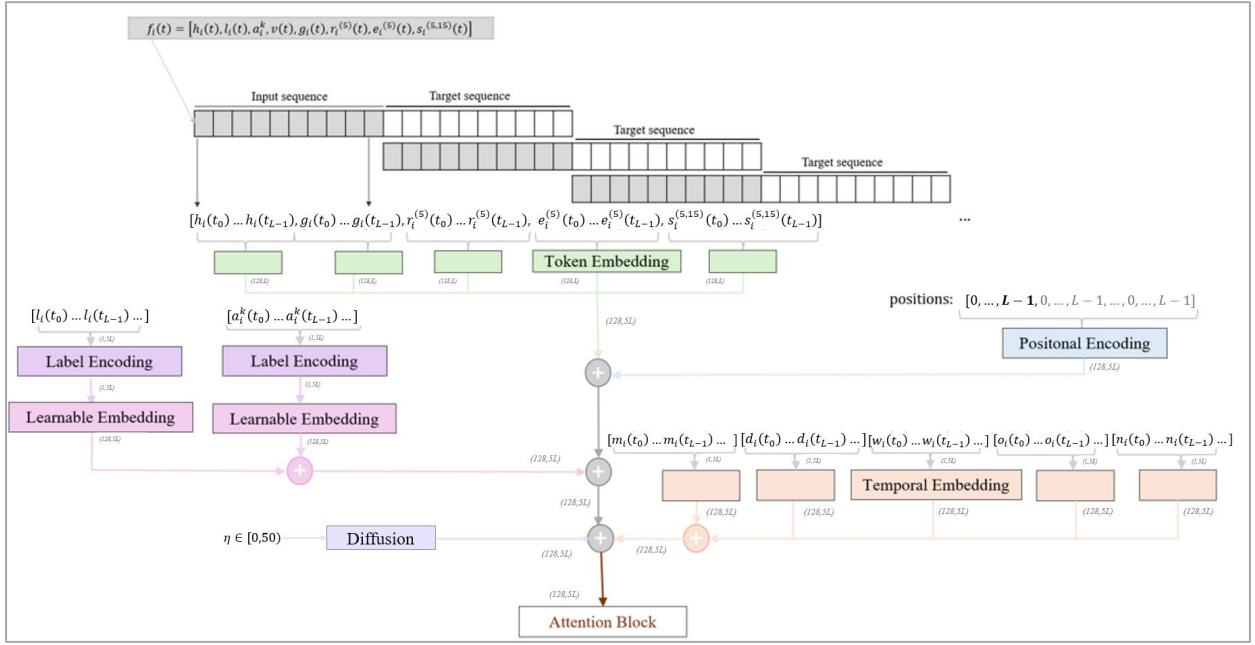

**Figure 2.** Encoding & Embedding structure

To accurately predict HR patterns during physical activities the model must recognize the temporal position of each time step, as HR changes throughout different activity stages. To achieve this, we create a position vector of length $5L$ (matching the input feature size), using repeating sequences of $L$ across five feature groups (HR, gradient, rolling standard deviation, exponential moving average, and trend). Each position is mapped to a ($d_{model}, L$) embedding, forming a ($d_{model}, 5L$) positional embedding tensor. This is added elementwise to the feature embedding tensor, allowing the model to learn both feature content and its position in the sequence, enhancing temporal pattern recognition. The context of each HR value is further enriched by incorporating information about the type of activity performed and the corresponding intensity level. Each of these features undergoes a similar processing pipeline. The categorical variables are first transformed into integer indices via label encoding and subsequently passed through learnable embedding layers based on dedicated embedding lookup tables. The embedding vector is not only learned, but also scaled dynamically by two trainable scalar parameters, α and β, as follows:

$$E_{activity} = \alpha \cdot Embedding(activity) + \beta \tag{25}$$

where $Embedding$ is a function that receives the categorical index corresponding to a specific physical activity or intensity level and outputs a vector representation. The learnable parameters $\alpha$ and $\beta$ allow the model to adaptively scale and shift the embeddings before integrating them into the final token representation. The resulting tensor has a shape of ($d_{model}, 5L$), consistent with the other embedding components. These two tensors, corresponding to the embedded activity level and intensity level, are summed elementwise with the previously constructed tensor that combines the embedded positional and numerical features. This unified representation integrates physiological and contextual information for each time step, allowing the model to learn HR dynamics in relation to both position and activity context. We incorporate a diffusion step embedding to represent the current step within the diffusion process. The embedding tensor is generated using sinusoidal positional encodings, followed by learned nonlinear transformations and is summed elementwise to the existing token representation.

Finally, the temporal features use the same procedure as with previous embeddings, ensuring that the temporal context is aligned with the rest of the input, providing a corresponding temporal reference for each token in the sequence. Each temporal component is first encoded as a categorical index and then mapped to an embedding vector using a fixed embedding approach via sinusoidal encoding. The final temporal embedding is obtained by summing the embeddings of all five temporal components, each of shape ($d_{model}, 5L$), into a single tensor of the same shape ($d_{model}, 5L$). This composite embedding allows the model to capture and leverage periodicities and temporal regularities across different time scales.

Once the input embeddings are constructed, they are passed through a series of attention blocks, where each token embedding is updated based on its relationship with all other tokens in the sequence. These attention-driven updates allow the model to dynamically integrate contextual information across time, such as temporal dependencies in HR fluctuations during different phases of physical activity. Following the attention layers, the updated embeddings are processed by feed-forward neural networks, where each token embedding is handled independently and in parallel through a series of linear and nonlinear transformations. The output is then added back to the original input from the attention block via a residual connection, enabling the model to retain both the contextualized representations of HR patterns and the nonlinear features learned by the feed-forward neural network.

Each target sequence $W_i^{target}(t)$, consists of a vector with L elements, where each element $f'_i(t)$ stores the HR and the activity performed at time t. This sequence is passed through a target embedding layer, which encodes the HR, activity and diffusion step information using the same method as the input embedding. Each embedded feature produces a tensor of shape ($d_{model}, L$), and these tensors are summed to capture the combined semantic representation of the target features. Afterward, positional information is added. The positional indices for the target sequence range from L to 2L−1, and are embedded using positional encoding, producing a tensor of shape ($d_{model}, L$). This positional embedding tensor is then added to the output embeddings to incorporate temporal context into the target representation.

After embedding, the target sequence enters the decoder layers. The decoding process starts with masked multi-head self-attention, which ensures that each position in the sequence can only attend to earlier positions. Following this, a cross-attention mechanism allows the decoder to attend to the encoder's outputs.

## 3.2. Activity-specific encoder and attention mechanism

In the original Transformer architecture, all encoder layers are generic, meaning that every input passes through the entire stack of encoder layers before being fed into the decoder via cross-attention. To enhance the model's ability to capture activity-specific patterns, we modified the last encoder layer to specialize based on the type of activity being performed. Therefore, we created specific encoders for each unique activity type and replaced the last encoder layer with them (Figure 3).

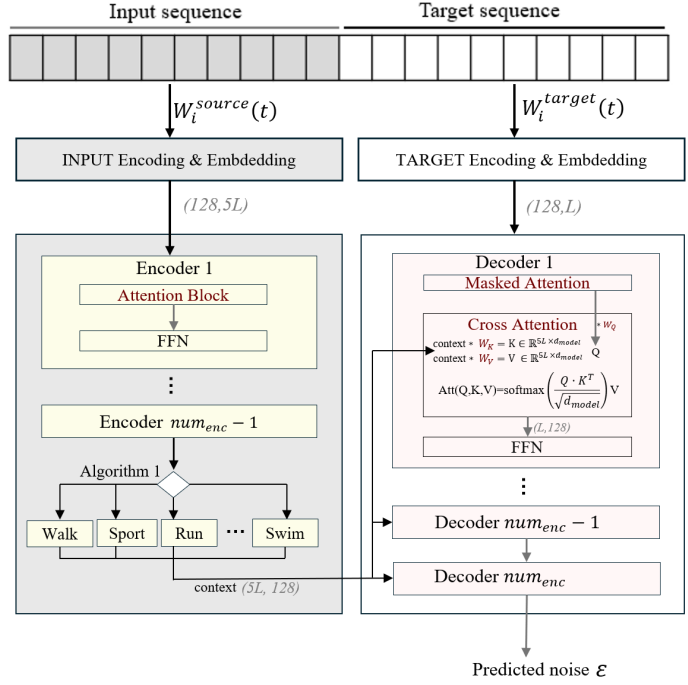

**Figure 31.** Specialized encoders based on activity type

The proposed Transformer model includes $num_{enc}$ - 1 generic encoder layers (lines 1–3 in Algorithm 1) shared across all input sequences, providing a common representation. The role of the generic encoders is to extract general HR patterns and relationships within the data. These layers focus on learning common dependencies, such as HR evolution over time across different intensities. The final encoder layer is activity-specific, meaning that different encoder instances are assigned to different activity types, allowing the model to specialize based on activity context (line 8). A redirection mechanism is introduced to extract all sequences from a batch and forward them to the specific encoder (lines 6-10). After all sequences have been individually processed by their respective specialized encoders, they are recombined to reconstruct the original batch structure. This merged output is then passed through a layer normalization step and subsequently used as input to each decoder layer (lines 13–14). This ensures that all decoder layers consistently attend to a unified representation of the input sequence.

Within the cross-attention mechanism, this normalized encoder output is used to generate the <Key, Val> tensors by multiplying it with dedicated learnable projection matrices, the key projection matrix and the value projection matrix, respectively. To compute the Query tensors, the model uses the output from the previous decoder layer, specifically, the result obtained after masked self-attention followed by Add & Norm. This decoder output is then multiplied by a separate learnable query projection matrix specific to the cross-attention layer, enabling the decoder to attend to relevant parts of the encoded input context during prediction.

In the cross-attention mechanism, the model computes attention for each of the L points in the output sequence with respect to the 5L points in the input sequence. This is done by calculating an attention score matrix as follows:

$$Scores = \frac{Q \cdot K^T}{\sqrt{d_{model}}} \quad (26)$$

where $Q \in \mathbb{R}^{L \times d_{model}}, K \in \mathbb{R}^{5L \times d_{model}} \rightarrow Scores \in \mathbb{R}^{L \times 5L}$

| ALGORITHM 1: Encoder Input Routing |
|---|
| **Inputs:** |
| ***src*** - input sequences of shape ***B x 5L x d<sub>model</sub>*** (***B*** –the batch size ***5L*** – the length of the input sequence ***d<sub>model</sub>*** – the model's dimension) |
| ***δ*** – a list of size ***B***, including activity labels corresponding to each sequence of batch |
| **sharedEncoders** – list of num_enc-1 generic Transformer encoders |
| **activityEncoders** – list of **card\|A\|** specialized activity encoders |
| **Outputs:** |
| ***context*** – final encoded representations of the input sequences |
| **Begin** |
| 1.   $foreach\ encoder\ in\ genericEncoders\ do$ <br> 2.      $src = encoder(src)$ <br> 3.   $endFor$ <br> 4.   $src_{specialized} = [\ ]$ <br> 5.   $foreach\ batch\ in\ src\ do$ <br> 6.      $foreach\ sequence, i\ in\ batch\ do$ <br> 7.         $activityIndex = δ[i]$ <br> 8.         $specializedEncoder = activityEncoders[activityIndex]$ <br> 9.         $s = specializedEncoder(sequence)$ <br> 10.      $src_{specialized}.append(s)$ <br> 11.      $endFor$ <br> 12.  $endFor$ <br> 13.  $context = LayerNorm(src_{specialized})$ <br> 14.  $return\ context$ <br> **End** |

A softmax is applied across each row over the 5L input positions, resulting in attention weights for each output token. These weights are then used to compute a weighted sum over the Value tensor $V \in \mathbb{R}^{5L \times d_{model}}$ producing an output of shape $(L, d_{model})$. This resulting tensor serves as the base representation for each position in the target sequence. It integrates contextual information from the encoder, selectively focused via attention. As such, it becomes the foundation for the model's final prediction at each time step.

## 3.3. Laplace diffusion process

To improve the HR prediction accuracy in case of activities changes that lead to sudden variations, we integrate a hybrid Denoising Diffusion Probabilistic Model (DDPM) [56] that utilizes Laplace noise. It consists of a forward diffusion process that gradually adds noise to the HR values of the target sequence, and a reverse diffusion process removes the noise step by step. During physical activities, HR sequences often contain sudden spikes or drops caused by changes in intensity or recovery periods. Laplace noise handles these sudden jumps better due to its heavy tails, allowing for more extreme values to be reconstructed. The Transformer model is trained to predict the added noise and recover the original sequence during the denoising process.

We define a sequence of latent variables $h_1, h_2, \ldots, h_S$, where S is the number of diffusion steps. The *forward diffusion process* can be written as a Markov chain:

$$q(h_1, \dots, h_S | h_0) := \prod_{s=1}^{S} q(h_s | h_{s-1}) \tag{27}$$

where $h_0$ is a sequence of HR values to be forecasted and each transition is expressed as:

$$q(h_s | h_{s-1}) := Laplace\left(h_s; \sqrt{1-\beta_s} \cdot h_{s-1}, \frac{\beta_s}{\sqrt{2}}\right) \tag{28}$$

and $\beta_s \in [0, 0.999]$ is the noise level that follows a cosine schedule [57].

We represent the *cumulative effect* of the forward process from the initial state $h_0$ to step $h_s$, by defining the cumulative product of all retention rates up to $s$ as:

$$\bar{\alpha}_s = \prod_{k=1}^{s} \alpha_k = \prod_{k=1}^{s} (1 - \beta_k) \tag{29}$$

where $\alpha_s = 1 - \beta_s$ is the fraction of the signal retained at step s.

The *marginal effect* $q(h_s | h_0)$ can be sampled as:

$$h_s = \sqrt{\bar{\alpha}_s} h_0 + \sqrt{1 - \bar{\alpha}_s} \cdot \varepsilon_s, \quad \varepsilon_s \sim Laplace(0, b_s) \tag{30}$$

where $\varepsilon_s$ is the Laplace-distributed noise with scale parameter $b_s = \sqrt{\frac{1-\bar{\alpha}_s}{2}}$. The scaling factor $\frac{1}{\sqrt{2}}$ is applied to match the noise variance of the Gaussian diffusion:

$$Var(\varepsilon_s) = 2b_s^2 = 2\left(\sqrt{\frac{1-\bar{\alpha}_s}{2}}\right)^2 = 1 - \bar{\alpha}_s \tag{31}$$

Following the Central Limit Theorem, the sum of scaled Laplace noises converges to a Gaussian.

The model is trained to learn the reverse denoising process by predicting the noise variable $\varepsilon_s$ used in the forward diffusion process. The *reverse diffusion process* is defined by a Markov chain as well:

$$p_\theta(h_{0:S}) := p(h_S) \prod_{s=1}^{S} p_\theta(h_{s-1} | h_s) \tag{32}$$

where each transition can be written as:

$$p_\theta(h_{s-1} | h_s) := Laplace\left(h_{s-1}; \mu_\theta(h_s, s), \frac{\tilde{\beta}_s}{\sqrt{2}}\right) \tag{33}$$

and $\mu_\theta(h_s, s)$ is derived from the forward process posterior:

$$\mu_\theta(h_s, s) = \frac{1}{\sqrt{\alpha_s}} \left( h_s - \frac{1 - \alpha_s}{\sqrt{1 - \bar{\alpha}_s}} \cdot \varepsilon_\theta(h_s, s) \right) \tag{34}$$

To maintain diversity during sampling, we add Laplace noise at each reverse step based on the time-dependent variance $\tilde{\beta}_s$ defined as:

$$\tilde{\beta}_s = \begin{cases} \frac{1-\alpha_{s-1}}{1-\alpha_s} \beta_s & s > 1 \\ \beta_1 & s = 1 \end{cases} \tag{35}$$

During training, a random diffusion step $s$ is selected for each sample. The raw HR sequence is then corrupted with Laplace noise $\varepsilon$ according to the predefined diffusion schedule at that step. The Transformer model is trained to predict this added noise $\varepsilon$, using the noisy sequence as input along with contextual embeddings and an embedding of the current diffusion step s. At inference time, the model is used to perform the reverse diffusion process. It starts with pure Laplace noise and runs for S denoising steps. At each step s, the Transformer receives the current noisy sequence $h_s$, along with the target-specific embeddings, and predicts the noise component $\varepsilon_s$. The predicted noise is then subtracted from the from $h_s$ (38) to obtain a cleaner version $h_{s-1}$. Repeating this process gradually reconstructs the final predicted HR sequence. Since each reverse process generates slightly different predictions due to the random noise added at each step, we generate multiple forecasts to improve robustness. The final predicted HR values are obtained by taking the median value across all generated samples at each time step.

## 4. Evaluation results

To evaluate the proposed transformer model, we have used a dataset comprising of an anonymized HR time series and the corresponding activity data of 29 unique persons. The data was collected using Fitbit fitness trackers over a period of 4 months. The dataset comprises approximately 20,000 HR measurements and about 294 labeled activity sessions, including detailed information on patterns, activity intensity levels, and sleep data (see Table 1).

**Table 1.** Activities from the dataset

| Activity Label | Description |
|---|---|
| Aerobic Workout | Includes aerobic activities with continuous movement |
| Outdoor Bike | Leisurely cycling - less than 16 km/h |
| Run | Running at 8 km/h |
| Sport | Includes high-intensity sports activities such as tennis, basketball and others. |
| Swim | Swimming at less than 23 meters/min |
| Walk | Walking less than 3 km/h, strolling very slowly |

The distribution of activities within the dataset reveals a predominance of low-to-moderate intensity exercises, such as walking and cycling, with fewer instances of high-intensity workouts. Table 2 presents a summary of key statistical measures, providing insights into user behavior across various physical activities. Walking and aerobic workouts exhibit the longest durations, whereas activities such as running, cycling, and swimming tend to be shorter due to their higher intensity. The HR distribution aligns with activity intensity, as dynamic activities demonstrate higher average and median Beats per Minute (BPM), along with increased calorie expenditure. These provide valuable insights into the relationship between activity type and physiological response, which are further correlated and contextualized.

**Table 2.** Activity metadata

| Activity Name | Total | % of Total | Avg. Duration (min) | Median HR (BPM) | Avg. Calories Burned |
|---|---|---|---|---|---|
| Walk | 175 | 59.5% | 89 | 99 | 274 |
| Run | 36 | 12.2% | 70 | 128 | 491 |
| Aerobic Workout | 34 | 10% | 86 | 112 | 736 |
| Outdoor Bike | 21 | 6.4% | 33 | 90 | 175 |
| Sport | 14 | 4.7% | 39 | 105 | 300 |
| Swim | 11 | 3.7% | 43 | 110 | 311 |
| Treadmill | 3 | 1% | 130 | 75 | 601 |

## 4.1. Data pre-processing

Data from wearables are prone to signal variations, missing values, and noise, which can affect model accuracy. Therefore, we used a preprocessing pipeline using preprocessing techniques such as normalization, feature selection, and missing data handling to improve data quality and consistency.

To address inconsistencies in HR readings, we employed smoothing techniques aimed at reducing short-term irregularities and enhancing the interpretability of temporal trends. As HR data tends to be autocorrelated, we evaluated several approaches, including the Moving Average, Random Walk, and Exponential Moving Average methods. Their effectiveness was assessed by computing the RMSE (Root Mean Squared Error) between the raw and smoothed datasets. The Moving Average applied on a window size of five data points, yielded the lowest RMSE value of 1.13, indicating superior noise reduction while preserving essential signal characteristics. Figure 4 illustrates the HR data before and after the smoothing process, highlighting the improvements in pattern clarity.

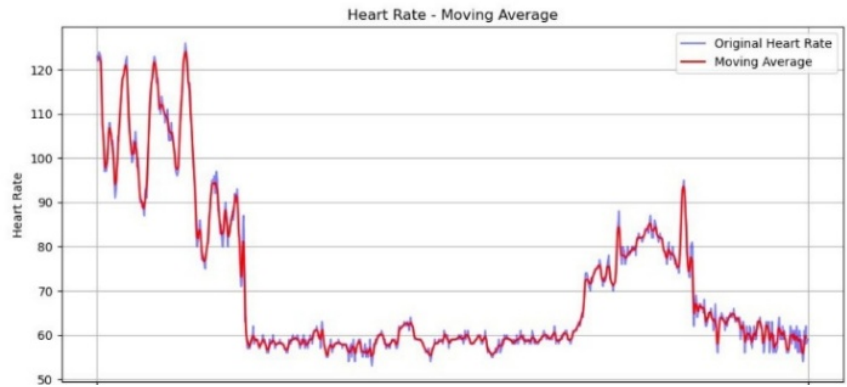

**Figure 4.** Smoothed HR data

Each structure $f_i(t + \xi)$ of the input sequence $W_i^{source}(t)$ contains a HR value and an intensity level at time $(t + \xi)$, where $\xi \epsilon \mathbb{N} \cap [0, L]$. This contextual information about intensity level is important for detecting outliers, because each intensity category is expected to correspond to a specific range of HR values. To further explore how outliers are distributed relative to intensity levels, we constructed a boxplot illustrating the HR distribution across each category (Figure 5). The Y-axis represents HR values, measured in BPM. Each box shows the interquartile range (IQR), containing the middle 50% of the data, with the central line indicating the median HR for that category. Outliers are represented as dots outside the whiskers, corresponding to values falling beyond 1.5 times the IQR.

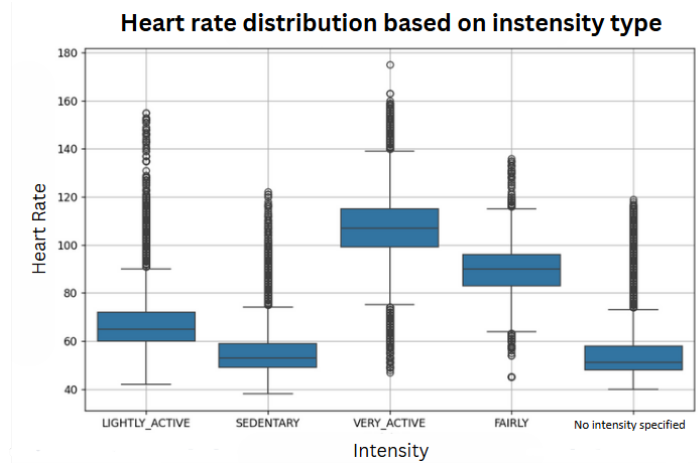

**Figure 5.** Intensity and HR boxplot

The boxplot reveals distinct trends where higher activity intensities correspond to elevated HR values. The "Very Active" category demonstrates the highest median HR and the broadest IQR, reflecting the expected variability during peak-intensity activities. The "Sedentary" category shows the lowest HR values but with a notable concentration of outliers, suggesting that some sudden spikes are more likely due to device errors or data recording issues rather than actual physiological responses. The "Fairly Active" and "Lightly Active" categories display overlapping HR ranges, which may reflect transitional periods between activity intensities or user-specific cardiovascular responses. To address the identified outliers, we used the Isolation Forest method [60] with a contamination rate of 0.05. Figure 6 illustrates the outcome of this method for a specific period, where red dots represent identified outliers, clearly separating them from normal HR fluctuation.

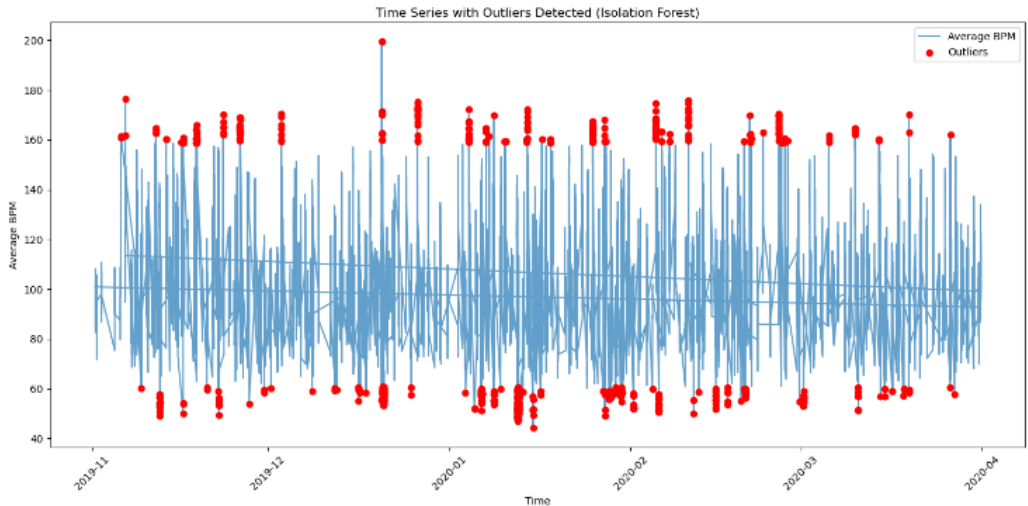

**Figure 6.** Detected outliers during activities

To analyze HR volatility during physical activities, we define sudden HR changes as transitions where the HR differences between the current measurement and any of the subsequent 1, 2, or 3 data points exceeds a threshold of 10 BPM. The distribution of change magnitudes (Figure 7a) has a mean of 13 BPM, and shows that, while changes around 10-14 BPM are the most common, larger shifts of 15-20 BPM occur regularly. Moreover, around 14% of total HR data points represent sudden jumps or drops, with high-intensity activities like running and aerobic workouts showing proportionally more sudden changes relative to their total measurements (Figure 7b). Figure 7c shows that these changes occur in both directions (rises and drops), with a slight bias towards increases (54.8%), which may indicate transitions into more intense activity phases. Activity sessions show varying intensity patterns (Figure 7d), which reflects how users adjust their effort levels during exercise. Treadmill, walking and outdoor bike activities, which often involve varying speeds or inclines, show the highest average number of intensities shifts per session. We analyzed HR variability by computing the average rolling standard deviation across 10-minute windows (Figure 7e). Dynamic activities like running and aerobic workout show a high variability (6.3 BPM and 5.2 BPM), while controlled indoor activities produce more stable patterns. These results demonstrate that sudden HR changes are common and systematic responses to activity modifications. Moreover, the HR responses during exercises are not only volatile but also structured and activity dependent. The frequency and magnitude of these sudden changes justify the use of heavy-tailed Laplace distributions in the diffusion process.

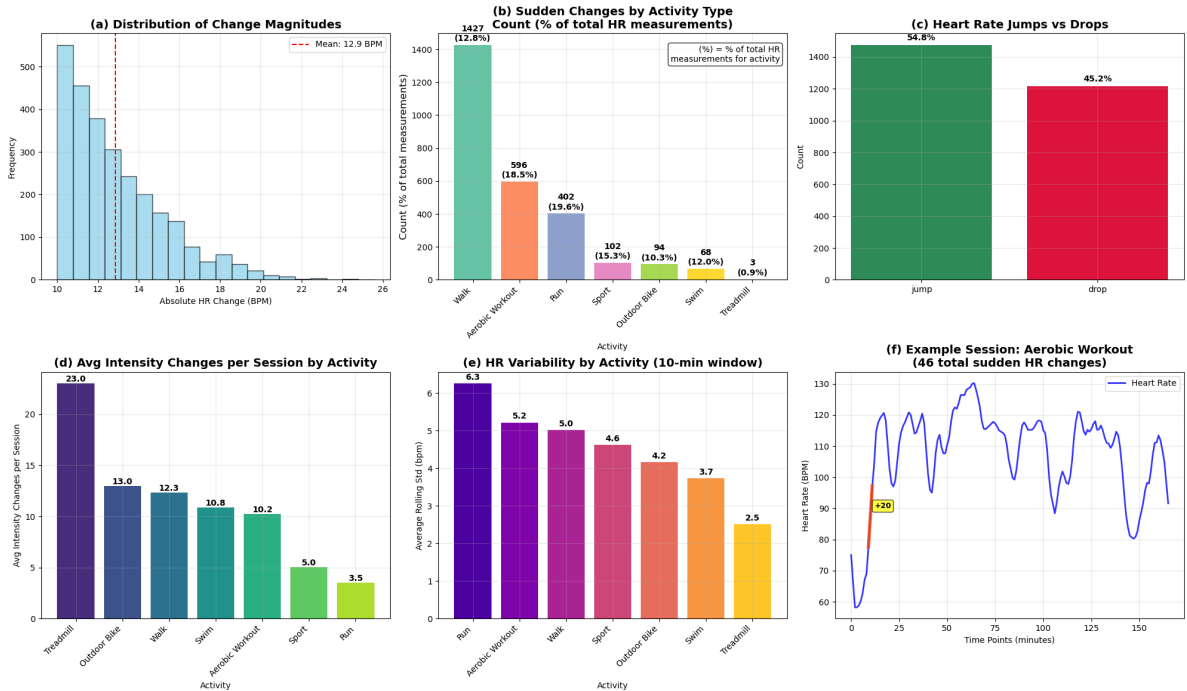

**Figure 7.** Analysis of sudden HR changes during physical activities

## 4.2. Performance evaluation

We assess the effectiveness of the proposed Transformer model by conducting a series of experiments to evaluate its accuracy and ability to capture activity-specific HR dynamics. Moreover, we compare our error metrics with other models designed for HR forecasting and discuss the results.

To evaluate the predictive power of the model, we report on four metrics that present main aspects of forecast accuracy (see Table 3 for aggregate results and Table 4 for activity-specific breakdowns).

**Table 3.** Error metrics values

| Metric | MAE | MAPE | RMSE | $R^2$ |
|---|---|---|---|---|
| Value | 2.19 | 2.3% | 3.44 | 0.97 |

MAE metric directly aligns with the training objective, represents average deviation from true HR, and it's interpretable in BPM. The Mean Absolute Percentage Error (MAPE) provides scale-independent interpretation, enabling comparison across activities with different HR ranges. We evaluate performance during sudden HR spikes (e.g., transitions between exercise intensities) using the RMSE. Finally, the Coefficient of Determination ($R^2$) shows how well the model explains variance in HR data, offering additional insights into the model's ability to understand the variability in HR signals.

**Table 4.** Error metrics values per activity

| Activity/Metric | MAE | MAPE | RMSE |
|---|---|---|---|
| Aerobic Workout | 1.88 | 1.79% | 2.66 |
| Outdoor Bike | 2.02 | 2.34% | 2.52 |
| Run | 2.75 | 2.09% | 4.60 |
| Sport | 3.72 | 4.44% | 5.15 |
| Swim | 2.04 | 1.80% | 2.43 |
| Walk | 2.16 | 2.46% | 3.48 |
| Treadmill | 1.54 | 1.93% | 1.89 |

Figure 8 shows the model's HR predictions (dashed lines) against ground truth values (solid lines) for different activity segments across 10-time steps. Subfigures (a), (b), and (d) represent walking sessions with varying intensity and trend, while (c) displays a more dynamic aerobic workout session. The predictions follow the HR patterns well, proving that the model can adapt to both gradual and abrupt transitions. In subfigure (c), the model tracks the initial decrease and the later rise in HR, which is typical to short rest periods during high-intensity activities. Both increases (d) and decreases (b) are accurately captured, as well as increases followed by slower declines (a).

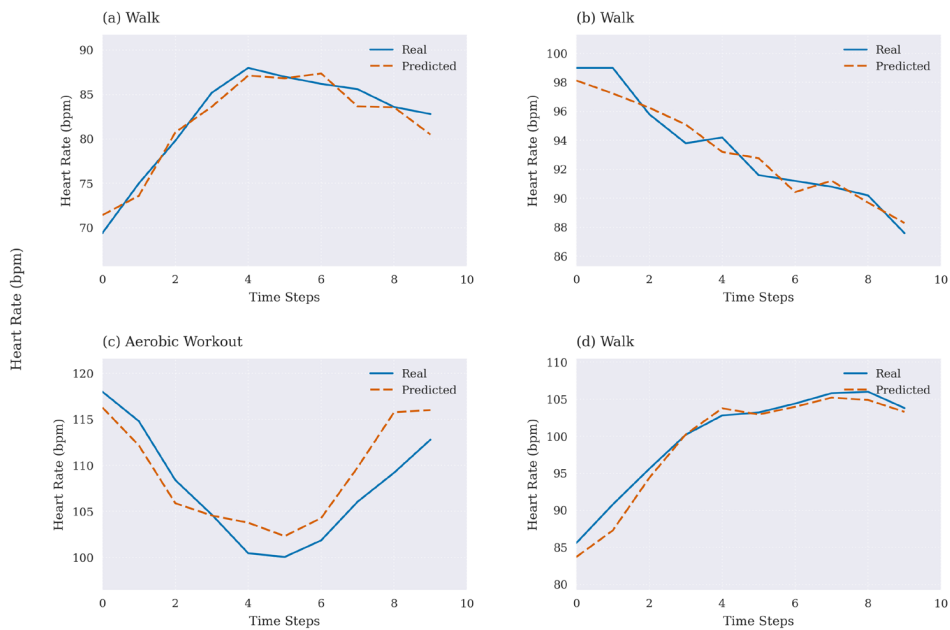

**Figure 8.** HR predictions across different activity segments

We compare the results of our paper against three baselines: vanilla transformer without activity conditioning or diffusion components, and two state of the art diffusion models for time series forecasting: CSDI [23] and TDSTF [55]. All models were trained on the same train/validation/test splits using the same preprocessing pipeline. In all three baseline models, we incorporated activity, intensity and temporal information as auxiliary features through standard embedding layers. The results are presented in Table 5.

**Table 5.** Comparative performance against baseline models for activity-conditioned HR forecasting

| Model | MAE | RMSE | $R^2$ |
|---|---|---|---|
| Vanilla Transformer | 6.86 | 10.19 | 0.77 |
| CSDI [23] | 5.09 | 6.89 | 0.83 |
| TDSTF [55] | 3.86 | 5.34 | 0.93 |
| Current approach | **2.19** | **3.44** | **0.97** |

Our proposed model achieves superior performance across all evaluation metrics, showing an improvement of 43% in MAE and 35% in RMSE compared to the best performing baseline (TDSTF). The vanilla Transformer shows poor performance, highlighting the importance of modeling the relationship between HR and activity context, rather than treating the activity information as general features. The diffusion baselines perform better than the vanilla Transformer, but struggle with activity transitions, suggested by the high RMSE values. The results further prove the effectiveness of the Laplace diffusion process to capture these transitions.

## 4.3. Discussion

We have conducted an analysis of the diffusion process parameters that determine how the model learns to reconstruct HR signals from noise. The two main hyperparameters evaluated were the noise schedule, which controls how noise is added during the forward diffusion process, and the number of diffusion steps, which affects both generation quality and computational efficiency. We evaluated three different scheduling approaches (Table 6): linear schedule, which provides uniform noise increase across timestamps; quadratic schedule, which introduces slower noise accumulation in early timestamps and accelerated noise addition in later stages; and cosine schedule, which implements a cosine-based noise progression that maintains signal details for extended periods before fast noise accumulation near the end of the forward process.

**Table 6.** Comparative performance of noise scheduling strategies

| Noise Schedule | MAE | RMSE | $R^2$ |
|---|---|---|---|
| Linear | 2.40 | 3.62 | 0.96 |
| Quadratic | 2.41 | 3.60 | 0.96 |
| Cosine | **2.19** | **3.44** | **0.97** |

The number of diffusion steps usually represents a trade-off between generation quality and computational efficiency. Using the optimal noise schedule identified above, we evaluated configurations with 50, 100 and 200 steps to explore their impact on the model's performance. The number of diffusion steps affects both the granularity of the denoising process and the computational cost during inference. The results show that the model achieves the best results with 50 diffusion steps, while maintaining the lowest inference time (Table 7).

**Table 7.** Comparative performance of total diffusion steps

| Diffusion steps | MAE | RMSE | $R^2$ |
|---|---|---|---|
| 50 | **2.19** | **3.44** | **0.97** |
| 100 | 2.99 | 4.54 | 0.94 |
| 200 | 3.66 | 6.36 | 0.88 |

The proposed model uses L1 loss, which corresponds to the negative log-likelihood of the Laplace distribution and enables learning the heavy-tailed characteristics needed for sharp HR transitions. To further validate this choice, we conducted comparative experiments to investigate whether the model would benefit from incorporating a small component of L2 loss for smaller residuals while maintaining L1 characteristics for larger errors, via Huber loss. We also evaluated DIALTE loss [58], designed for temporal sequence modeling. We present the experimental results in Table 8 confirming that L1 loss provides optimal performance for our diffusion process.

**Table 8.** Comparative performance of different loss functions

| Loss Function | Parameters | MAE | RMSE | $R^2$ |
|---|---|---|---|---|
| L1 | | **2.19** | **3.44** | **0.97** |
| Huber | $\delta = 1$ | 2.93 | 4.39 | 0.94 |
| Huber | $\delta = 0.4$ | 2.61 | 4.19 | 0.95 |
| Huber | $\delta = 0.1$ | 2.57 | 4.08 | 0.95 |
| DIALTE | $\alpha = 0.75$ | 3.85 | 4.60 | 0.92 |

We employed the Optuna framework [61] for automated optimization of all remaining model hyperparameters, including model dimensions, attention heads, learning rate, batch size, dropout rate and training epochs. To increase the number of encoder and decoder layers while maintaining training stability, we utilized a skip connection mechanism that collects intermediate representations from all transformer blocks to a convolutional decoder. The data was divided into 3 sets: training (65%), validation (15%), testing (20%). The Adam optimizer [62] was used to adjust model parameters during training, with a defined learning rate and a weight decay term for regularization. A learning rate scheduler is used to reduce the learning rate at predefined milestones to ensure easy convergence. An early stopping mechanism is used to prevent overfitting. The final configuration is presented in Table 9.

**Table 9.** Final model configuration

| Hyperparameter | Value |
| --- | --- |
| Embedding dimension | 128 |
| Generic encoders | 1 |
| Decoders | 2 |
| Transformer blocks | 3 |
| Attention heads | 8 |
| Learning rate | 0.001 |
| Dropout | 0.1 |
| Batch size | 32 |
| Training epochs | 400 |

# 5. Conclusion

In this paper, we developed an activity-conditioned Transformer architecture integrated with a Laplace diffusion process to address the challenges of HR modeling and forecasting during physical activities. We introduced activity-contextualized embeddings that encode categorical activity types, ordinal intensity levels, and continuous HR-derived features into a unified representation. The model incorporates activity-specific encoder layers that learn physiological patterns unique to different exercises, treating activity information as core components. Additionally, we integrated a Laplace diffusion process to capture abrupt HR transitions during activity changes and recovery phases. The heavy-tailed properties of Laplace distribution can effectively reconstruct sharp HR spikes and drops, which are characteristic of exercise transitions. The model was validated using a private real-world dataset collected from 29 patients over a 4-month period. The experimental results demonstrate our model outperforms current state-of-the-art methods, achieving a 43% improvement in MAE over the best baseline. The coefficient of determination $R^2$ is 0.97 which demonstrates strong alignment between predicted and actual HR values. These results confirm that the proposed model can serve as a practical tool for both healthcare providers and RPM systems. The model is currently integrated into the TransCare project RPM platform [59] and will be further validated in large scale trials defined within the project.

As future work, we plan to integrate additional physiological features into our model, utilizing the flexible embedding architecture to incorporate other vital signs such as blood pressure, oxygen saturation levels, and respiratory rate. Furthermore, we aim to explore the integration of embeddings generated by specialized healthcare LLMs that have been pre-trained on extensive clinical data. Recent advancements in healthcare LLMs have demonstrated their ability to encode rich, context-aware representations of physiological and clinical data that standard feature engineering methods might miss. Additionally, we plan to explore Denoising Diffusion Implicit Models (DDIM) sampling strategies to improve computational efficiency during inference while maintaining prediction quality, making the model more suitable for real-time monitoring applications. Finally, the model effectiveness will be assessed by healthcare professionals, whose valuable feedback can drive future refinement and optimization.

**Acknowledgement**

This work was supported by a grant of the Ministry of Research, Innovation and Digitization, CNCS/CCCDI - UEFISCDI, project number ERANET-PARTENERIAT-THCS-TransCare-1, within PNCDI IV.